\documentclass[10pt,twocolumn,letterpaper]{article}

\usepackage{iccv}
\usepackage{times}
\usepackage{epsfig}
\usepackage{graphicx}
\usepackage{amsmath}
\usepackage{amssymb}

\usepackage{algorithm}
\usepackage{algorithmic}

\usepackage{multirow}
\usepackage{array}
\usepackage{rotating}

\usepackage[pagebackref=true,breaklinks=true,letterpaper=true,colorlinks,bookmarks=false]{hyperref}

\iccvfinalcopy 
\hypersetup{
    bookmarks=true,         
    unicode=false,          
    pdftoolbar=true,        
    pdfmenubar=true,        
    pdffitwindow=false,     
    pdftitle={Certificate},    
    pdfauthor={Zhaowen Wang},     
    pdfsubject={},   
    pdfcreator={Zhaowen Wang},   
    pdfproducer={},  
    pdfkeywords={}, 
    pdfnewwindow=true,      
    colorlinks=false,       
    linkcolor=red,          
    citecolor=green,        
    filecolor=magenta,      
    urlcolor=cyan           
}

\iccvfinalcopy 

\def\iccvPaperID{526} 

\renewcommand{\mathbf}{\boldsymbol}
\newcommand{\y}{\boldsymbol y}

\newcommand{\x}{\boldsymbol x}

\newcommand{\D}{\boldsymbol D}

\newcommand{\aalpha}{\boldsymbol \alpha}
\newcommand{\z}{\boldsymbol z}

\ificcvfinal\fi
\begin{document}

\title{Deep Networks for Image Super-Resolution with Sparse Prior}


\author{
Zhaowen Wang\textsuperscript{\dag}\textsuperscript{\ddag} \; Ding Liu\textsuperscript{\dag} \; Jianchao Yang\textsuperscript{\S} \;
Wei Han\textsuperscript{\dag} \; Thomas Huang\textsuperscript{\dag}\\
\textsuperscript{\dag}Beckman Institute, University of Illinois at Urbana-Champaign, Urbana, IL \\
\textsuperscript{\ddag}Adobe Research, San Jose, CA \;\; \textsuperscript{\S}Snapchat, Venice, CA\\
{\tt\small \textsuperscript{\ddag}zhawang@adobe.com \; \textsuperscript{\dag}\{dingliu2, weihan3, huang\}@ifp.uiuc.edu \textsuperscript{\S}jcyangenator@gmail.com}
}

\maketitle

\begin{abstract}
Deep learning techniques have been successfully applied in many areas of computer vision,
including low-level image restoration problems.
For image super-resolution, several models based on deep neural networks have been recently
proposed and attained superior performance that overshadows all previous handcrafted models.
The question then arises whether large-capacity and data-driven models have
become the dominant solution to the ill-posed super-resolution problem.
In this paper, we argue that domain expertise represented by the conventional sparse coding model
is still valuable, and it can be combined with the key ingredients of deep learning to achieve further
improved results.
We show that a sparse coding model particularly designed for super-resolution
can be incarnated as a neural network, and trained in a cascaded structure from end to end.
The interpretation of the network based on sparse coding leads to
much more efficient and effective training, as well as a reduced model size.
Our model is evaluated on a wide range of images, and shows clear advantage over existing state-of-the-art methods
in terms of both restoration accuracy and human subjective quality.
\end{abstract}


\section{Introduction}
\label{sec:intro}

Single image super-resolution (SR) aims at obtaining a high-resolution (HR) image
from a low-resolution (LR) input image by inferring all the missing high frequency contents.
With the known variables in LR images greatly outnumbered by the unknowns in HR images,
SR is a highly ill-posed problem and the current techniques are far from being satisfactory for
many real applications \cite{Baker02PAMI,Lin04PAMI}.

To regularize the solution of SR, people have exploited various priors of natural images.
Analytical priors, such as bicubic interpolation, work well for smooth regions;
while image models based on statistics of edges \cite{Fattal07TG} and gradients \cite{Kim10PAMI,Aly05TIP}
can recover sharper structures.
In the patch-based SR methods, HR patch candidates are represented as the sparse linear combination of
dictionary atoms trained from external databases \cite{Yang10ScSR,Yang12coupled},
or recovered from similar examples in the LR image itself at different locations and across different scales \cite{Glasner09ICCV,Freedman11TG,wang2015learning}.
A comprehensive review of more SR methods can be found in \cite{Yang14ECCV}.

More recently, inspired by the great success achieved by deep learning \cite{Krizhevsky12NIPS,Sermanet13Overfeat,Vincent08ICML}
in other computer vision tasks, people begin to use neural networks with deep architecture for image SR.
Multiple layers of collaborative auto-encoders are stacked together in \cite{Cui14ECCV} for robust matching
of self-similar patches.
Deep convolutional neural networks (CNN) \cite{Dong14ECCV} and deconvolutional networks \cite{osendorfer2014image}
are designed that directly learn the non-linear mapping
from LR space to HR space in a way similar to coupled sparse coding \cite{Yang12coupled}.
As these deep networks allow end-to-end training of all the model components between LR input and HR output,
significant improvements have been observed over their shadow counterparts.

The networks in \cite{Cui14ECCV,Dong14ECCV} are built with generic architectures, which means
all their knowledge about SR is learned from training data.
On the other hand, people's domain expertise for the SR problem, such as natural image prior and image degradation model,
is largely ignored in deep learning based approaches.
It is then worthy to investigate whether domain expertise can be used to design better deep model architectures,
or whether deep learning can be leveraged to improve the quality of handcrafted models.

In this paper, we extend the conventional sparse coding model \cite{Yang10ScSR} using several key ideas from deep learning,
and show that domain expertise is complementary to large learning capacity in further improving SR performance.
First, based on the learned iterative shrinkage and thresholding algorithm (LISTA) \cite{Gregor10ICML},
we implement a feed-forward neural network whose layers strictly correspond to each step in the processing flow of sparse coding based image SR.
In this way, the sparse representation prior is effectively encoded in our network structure;
at the same time, all the components of sparse coding can be trained jointly through back-propagation.
This simple model, which is named sparse coding based network (SCN),
achieves notable improvement over the generic CNN model \cite{Dong14ECCV}
in terms of both recovery accuracy and human perception, and yet has a compact model size.
Moreover, with the correct understanding of each layer's physical meaning,
we have a more principled way to initialize the parameters of SCN, which helps to improve optimization speed and quality.

A single network is only able to perform image SR by a particular scaling factor.
In \cite{Dong14ECCV}, different networks are trained for different scaling factors.
In this paper, we also propose a cascade of multiple SCNs to achieve SR for arbitrary factors.
This simple approach, motivated by the self-similarity based SR approach \cite{Glasner09ICCV},
not only increases the scaling flexibility of our model,
but also reduces artifacts for large scaling factors.
The cascade of SCNs (CSCN) can also benefit from the end-to-end training of deep network
with a specially designed multi-scale cost function.

In short, the contributions of this paper include:
\vspace{-5pt}
\begin{itemize}
\setlength{\itemsep}{-2pt}
  \item combine the domain expertise of sparse coding and the merits of deep learning to achieve
        better SR performance with faster training and smaller model size;
  \item use network cascading for large and arbitrary scaling factors;
  \item conduct a subjective evaluation on several recent state-of-the-art methods.
\end{itemize}

In the following, we will first review related work in Sec.~\ref{sec:related}.
The SCN and CSCN models are introduced in Sec.~\ref{sec:rnn} and Sec.~\ref{sec:crnn},
with implementation details in Sec.~\ref{sec:exp_imp}.
Extensive experimental results are reported in Sec.~\ref{sec:exp},
and conclusions are drawn in Sec.~\ref{sec:cncl}.


\section{Related Work}
\label{sec:related}

\subsection{Image SR Using Sparse Coding}

The sparse representation based SR method \cite{Yang10ScSR} models the transform from
each local patch $\y\in\mathbb{R}^{m_y}$ in the bicubic-upscaled LR image to the corresponding patch $\x \in \mathbb{R}^{m_x}$ in the HR image.
The dimension $m_y$ is not necessarily the same as $m_x$ when image features other than raw pixel is used to represent patch $\y$.
It is assumed that the LR(HR) patch $\y$($\x$) can be represented with respect to
an overcomplete dictionary $\D_y$($\D_x$) using some sparse linear coefficients $\aalpha_y$($\aalpha_x$) $\in \mathbb{R}^{n}$,
which are known as sparse code.
Since the degradation process from $\x$ to $\y$ is nearly linear, the patch pair can share the same sparse code
$\aalpha_y=\aalpha_x=\aalpha$ if the dictionaries $\D_y$ and $\D_x$ are defined properly.
Therefore, for an input LR patch $\y$, the HR patch can be recovered as
\begin{equation}
\label{eq:scsr}
    \x = \D_x \aalpha, \;\; {\rm s.t.}\; \aalpha = \arg\min_{\z} \| \y - \D_y\z \|_2^2 + \lambda \| \z \|_1,
\end{equation}
where $\|\cdot\|_1$ denotes the $\ell_1$ norm which is convex and sparsity-inducing,
and $\lambda$ is a regularization coefficient.
The dictionary pair $(\D_y, \D_x)$ can be learned alternatively with the inference of training patches' sparse codes
in their joint space \cite{Yang10ScSR} or through bi-level optimization \cite{Yang12coupled}.

\subsection{Network Implementation of Sparse Coding}

\begin{figure}[t]
\centering
	\includegraphics[width=1.0\linewidth]{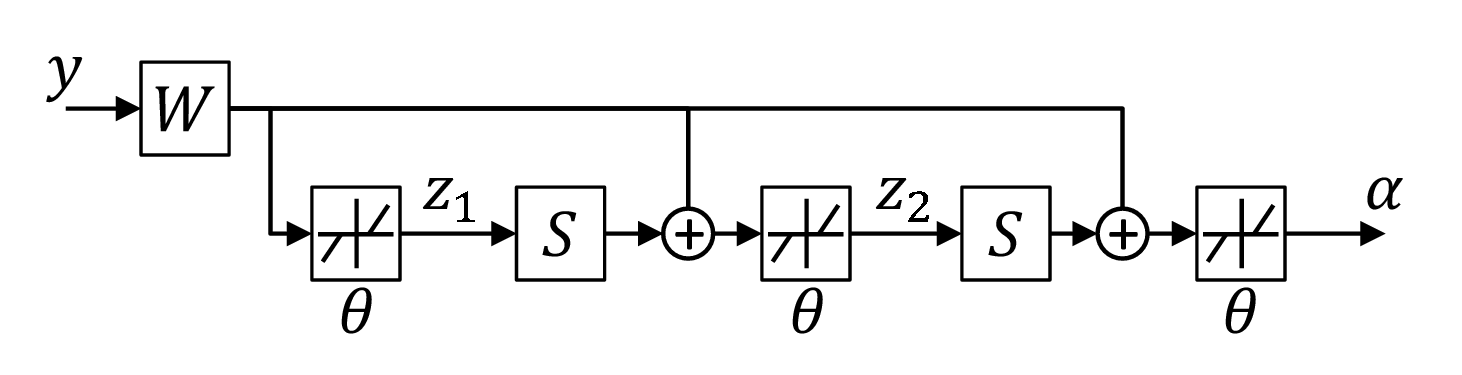}
\caption{A LISTA network \cite{Gregor10ICML} with 2 time-unfolded recurrent stages, whose output $\aalpha$ is an
        approximation of the sparse code of input signal $\y$. The linear weights $\mathbf{W}$, $\mathbf{S}$
        and the shrinkage thresholds $\mathbf{\theta}$ are learned from data.}
\label{fig:lista}
\end{figure}

There is an intimate connection between sparse coding and neural network, which has been well studied in \cite{kavukcuoglu10Fast,Gregor10ICML}.
A feed-forward neural network as illustrated in Fig.~\ref{fig:lista}
is proposed in \cite{Gregor10ICML} to efficiently approximate the sparse code $\aalpha$ of input signal $\y$
as it would be obtained by solving \eqref{eq:scsr} for a given dictionary $\D_y$.
The network has a finite number of recurrent stages, each of which updates the intermediate sparse code according to
\begin{equation}
    \z_{k+1} = h_{\mathbf{\theta}}(\mathbf{W}\y+\mathbf{S}\z_{k}) ,
\end{equation}
where $h_{\mathbf{\theta}}$ is an element-wise shrinkage function defined as
$[h_{\mathbf{\theta}}(\mathbf{a})]_i=\mathrm{sign}(a_i)(|a_i|-\theta_i)_+$
with positive thresholds $\mathbf{\theta}$.

Different from the iterative shrinkage and thresholding algorithm (ISTA) \cite{Daubechies04,Rozell08}
which finds an analytical relationship between network parameters (weights $\mathbf{W}$, $\mathbf{S}$ and thresholds $\mathbf{\theta}$)
and sparse coding parameters ($\D_y$ and $\lambda$), the authors of \cite{Gregor10ICML} learn
all the network parameters from training data using a back-propagation algorithm called learned ISTA (LISTA).
In this way, a good approximation of the underlying sparse code can be obtained within a fixed number of recurrent stages.

\section{Sparse Coding based Network for Image SR}
\label{sec:rnn}

\begin{figure*}[t]
\centering
    \hspace{100mm} 
	\includegraphics[width=0.8\linewidth]{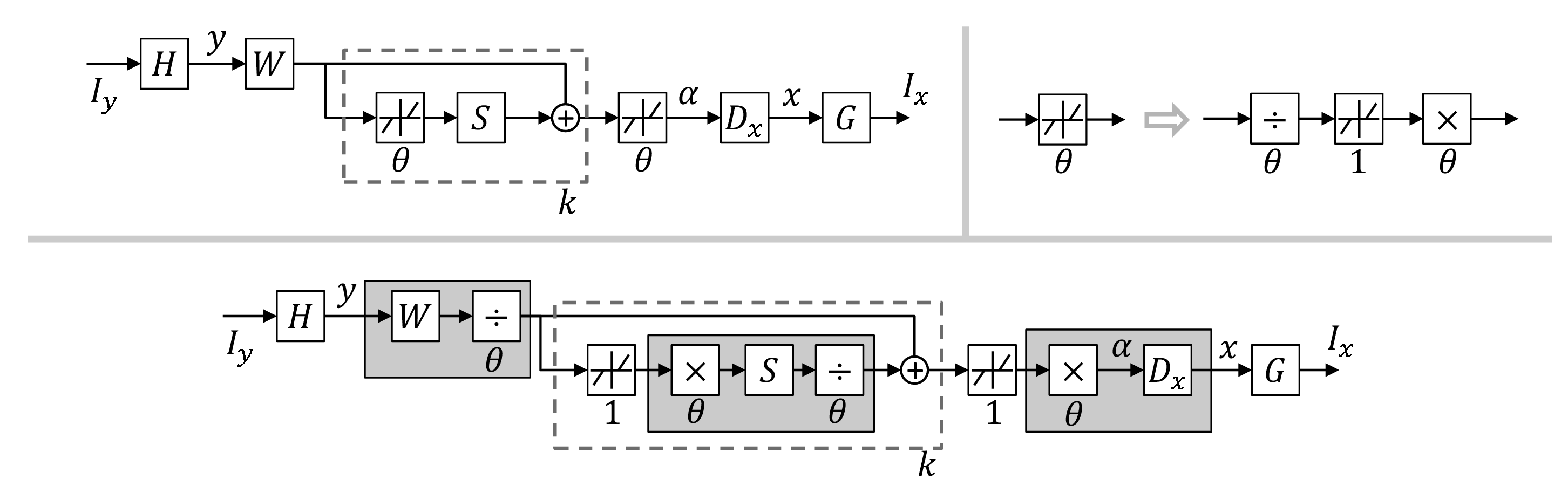}

\caption{Top left: the proposed SCN model with a patch extraction layer $\mathbf{H}$, a LISTA sub-network for sparse coding
         (with $k$ recurrent stages denoted by the dashed box), a HR patch recovery layer $\D_x$, and a patch combination layer $\mathbf{G}$.
         Top right: a neuron with an adjustable threshold decomposed into two linear scaling layers and a unit-threshold neuron.
         Bottom: the SCN re-organized with unit-threshold neurons and adjacent linear layers merged together in the gray boxes.}
\label{fig:rnn}
\end{figure*}

Given the fact that sparse coding can be effectively implemented with a LISTA network, it is straightforward to
build a multi-layer neural network that mimics the processing flow of the sparse coding based SR method \cite{Yang10ScSR}.
Same as most patch-based SR methods, our sparse coding based network (SCN) takes the bicubic-upscaled LR image $\mathbf{I}_y$ as input,
and outputs the full HR image $\mathbf{I}_x$.
Fig.~\ref{fig:rnn} shows the main network structure, and each of the layers is described in the following.

The input image $\mathbf{I}_y$ first goes through a convolutional layer $\mathbf{H}$ which extracts feature
for each LR patch. There are $m_y$ filters of spatial size $s_y{\times}s_y$ in this layer, so that our input patch size is $s_y{\times}s_y$
and its feature representation $\y$ has $m_y$ dimensions.

Each LR patch $\y$ is then fed into a LISTA network with a finite number of $k$ recurrent stages to obtain its sparse code $\aalpha \in \mathbb{R}^n$.
Each stage of LISTA consists of two linear layers parameterized by $\mathbf{W} \in \mathbb{R}^{n{\times}m_y}$ and $\mathbf{S} \in \mathbb{R}^{n{\times}n}$,
and a nonlinear neuron layer with activation function $h_{\mathbf{\theta}}$.
The activation thresholds $\mathbf{\theta}\in\mathbb{R}^n$ are also to be updated during training, which complicates the learning algorithm.
To restrict all the tunable parameters in our linear layers, we do a simple trick to rewrite the activation function as
\begin{equation}
\label{eq:unitact}
    [h_{\mathbf{\theta}}(\mathbf{a})]_i = \mathrm{sign}(a_i)\theta_i(|a_i|/\theta_i-1)_+ = \theta_i h_{1}(a_i/\theta_i).
\end{equation}
Eq.~\eqref{eq:unitact} indicates the original neuron with an adjustable threshold can be decomposed into two linear scaling layers
and a unit-threshold neuron, as shown in the top-right of Fig.~\ref{fig:rnn}. The weights of the two scaling layers
are diagonal matrices defined by $\mathbf{\theta}$ and its element-wise reciprocal, respectively.

The sparse code $\aalpha$ is then multiplied with HR dictionary $\D_x \in \mathbb{R}^{m_x{\times}n}$ in the next linear layer,
reconstructing HR patch $\x$ of size $s_x{\times}s_x=m_x$.

In the final layer $\mathbf{G}$, all the recovered patches are put back to the corresponding positions in the HR image $\mathbf{I}_x$.
This is realized via a convolutional filter of $m_x$ channels with spatial size $s_g{\times}s_g$.
The size $s_g$ is determined as the number of neighboring patches that overlap with the same pixel in each spatial direction.
The filter will assign appropriate weights to the overlapped recoveries from different patches and
take their weighted average as the final prediction in $\mathbf{I}_x$.

As illustrated in the bottom of Fig.~\ref{fig:rnn}, after some simple reorganizations of the layer connections,
the network described above has some adjacent linear layers which can be merged into a single layer.
This helps to reduce the computation load as well as redundant parameters in the network.
The layers $\mathbf{H}$ and $\mathbf{G}$ are not merged because we apply additional nonlinear normalization operations
on patches $\y$ and $\x$, which will be detailed in Sec.~\ref{sec:exp_imp}.

Thus, there are totally 5 trainable layers in our network: 2 convolutional layers $\mathbf{H}$ and $\mathbf{G}$,
and 3 linear layers shown as gray boxes in Fig.~\ref{fig:rnn}.
The $k$ recurrent layers share the same weights and are therefore conceptually regarded as one.
Note that all the linear layers are actually implemented as convolutional layers applied on each patch
with filter spatial size of $1{\times}1$, a structure similar to the network in network \cite{Lin13NIN}.
Also note that all these layers have only weights but no biases (zero biases).

Mean square error (MSE) is employed as the cost function to train the network,
and our optimization objective can be expressed as
\begin{equation}
\label{eq:objrnn}
    \min\limits_{\mathbf{\Theta}} \sum_i \|SCN(\mathbf{I}_y^{(i)}; \mathbf{\Theta}) - \mathbf{I}_x^{(i)} \|^2_2,
\end{equation}
where $\mathbf{I}_y^{(i)}$ and $\mathbf{I}_x^{(i)}$ are the $i$-th pair of LR/HR training data,
and $SCN(\mathbf{I}_y; \mathbf{\Theta})$ denotes the HR image for $\mathbf{I}_y$ predicted using
the SCN model with parameter set $\mathbf{\Theta}$.
All the parameters are optimized through the standard back-propagation algorithm.
Although it is possible to use other cost terms that are more correlated with human visual perception
than MSE, our experimental results show that simply minimizing MSE leads to
improvement in subjective quality.

\subsection*{Advantages over Previous Models}

The construction of our SCN follows exactly each step in the sparse coding based SR method \cite{Yang10ScSR}.
If the network parameters are set according to the dictionaries learned in \cite{Yang10ScSR}, it can reproduce almost the same results.
However, after training, SCN learns a more complex regression function and
can no longer be converted to an equivalent sparse coding model.
The advantage of SCN comes from its ability to jointly optimize all the layer parameters from end to end;
while in \cite{Yang10ScSR} some variables are manually designed and some
are optimized individually by fixing all the others.


Technically, our network is also a CNN and it has similar layers as the CNN model proposed in \cite{Dong14ECCV}
for patch extraction and reconstruction.
The key difference is that we have a LISTA sub-network specifically designed to enforce sparse representation prior;
while in \cite{Dong14ECCV} a generic rectified linear unit (ReLU) \cite{Nair10relu} is used for nonlinear mapping.
Since SCN is designed based on our domain knowledge in sparse coding,
we are able to obtain a better interpretation of the filter responses
and have a better way to initialize the filter parameters in training.
We will see in the experiments that all these contribute to better SR results, faster training speed and smaller model size
than a vanilla CNN.

\section{Network Cascade for Scalable SR}
\label{sec:crnn}

Like most SR models learned from external training examples, the SCN discussed previously
can only upscale images by a fixed factor.
A separate model needs to be trained for each scaling factor to achieve the best performance,
which limits the flexibility and scalability in practical use.
One way to overcome this difficulty is to repeatedly enlarge the image by a fixed scale until
the resulting HR image reaches a desired size.
This practice is commonly adopted in the self-similarity based methods \cite{Glasner09ICCV,Freedman11TG,Cui14ECCV},
but is not so popular in other cases for the fear of error accumulation
during repetitive upscaling.

\begin{figure}[t]
\small
\center
    \includegraphics[width=0.12\linewidth,angle=0,clip]{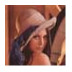} \\ (a) LR image
\begin{tabular}{cc}
    \includegraphics[width=0.45\linewidth,angle=0,clip]{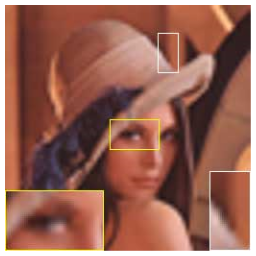} &
    \includegraphics[width=0.45\linewidth,angle=0,clip]{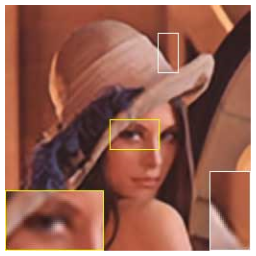} \\
    (b) bicubic$\times4$ (28.52) & (c) SCN$\times2$ \& bicubic$\times2$ (30.27) \\
    \includegraphics[width=0.45\linewidth,angle=0,clip]{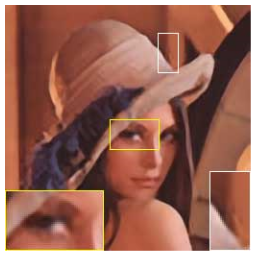} &
    \includegraphics[width=0.45\linewidth,angle=0,clip]{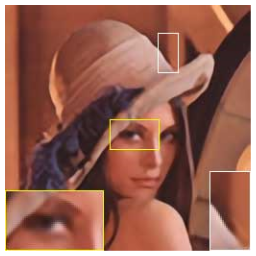} \\
    (d) SCN$\times4$ (30.22) & (e) SCN$\times2$ \& SCN$\times2$ (30.72) \\
\end{tabular}
\caption{SR results for the ``Lena'' image upscaled by 4 times. (a) $\rightarrow$ (b) $\rightarrow$ (d) represents
         the processing flow with a single SCN$\times4$ model. (a) $\rightarrow$ (c) $\rightarrow$ (e) represents
         the processing flow with two cascaded SCN$\times2$ models. PSNR is given in parentheses.}
\label{fig:lenacascade}
\end{figure}

In our case, however, it is observed that a cascade of SCNs (CSCN) trained for small scaling factors
can generate even better SR results than a single SCN trained for a large scaling factor,
especially when the target scaling factor is large (greater than 2).
This is illustrated by the example in Fig.~\ref{fig:lenacascade}.
Here an input image is magnified by $\times$4 times in two ways:
with a single SCN$\times$4 model through the processing flow (a) $\rightarrow$ (b) $\rightarrow$ (d);
and with a cascade of two SCN$\times$2 models through (a) $\rightarrow$ (c) $\rightarrow$ (e).
It can be seen that the input to the second cascaded SCN$\times$2 in (c) is already sharper and contains less artifacts
than the bicubic$\times$4 input to the single SCN$\times$4 in (b), which naturally leads to the better final
result in (e) than the one in (d).
Therefore, each SCN in the cascade serves as a ``relaying station'' which progressively recovers some useful information
lost in bicubic interpolation and compensates for the distortion aggregated from previous stages.


\begin{figure}[t]
\center
    \includegraphics[width=1.0\linewidth,angle=0,clip]{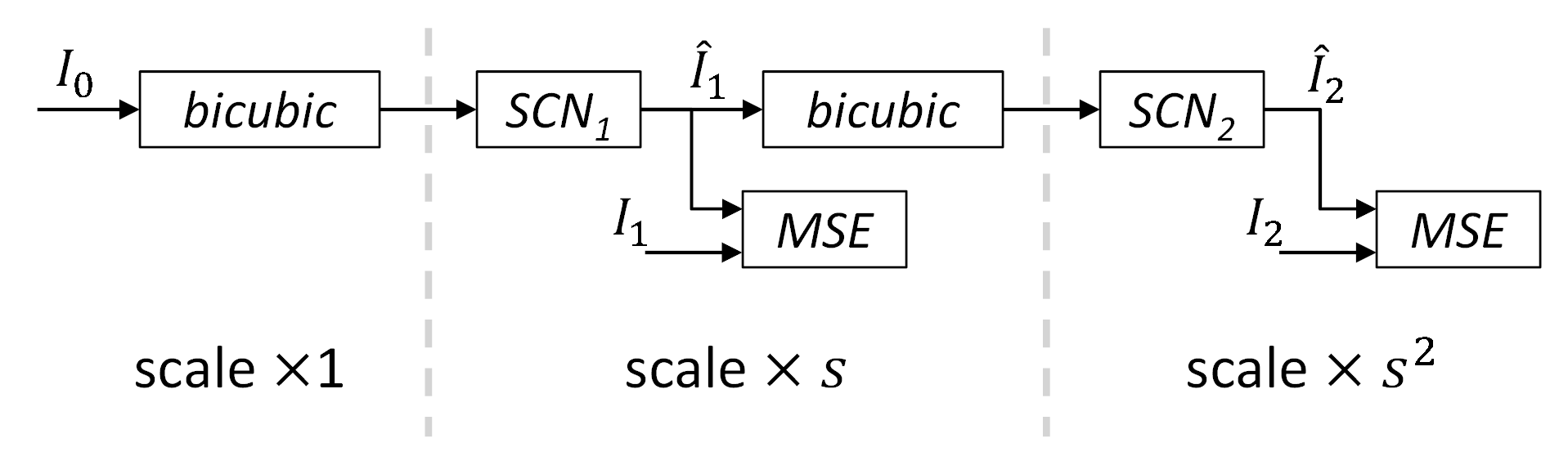}
\caption{Training cascade of SCNs with multi-scale objectives.}
\vspace{-1mm}
\label{fig:crnn}
\end{figure}

The CSCN is also a deep network, in which the output of each SCN is connected to
the input of the next SCN with bicubic interpolation in the between.
To construct the cascade, besides stacking several SCNs trained individually with respect to \eqref{eq:objrnn},
we can also optimize all of them jointly as shown in Fig.~\ref{fig:crnn}.
Without loss of generality, we assume each SCN in the cascade has the same scaling factor $s$.
Let $\mathbf{I}_0$ denote the input image of original size, and $\mathbf{\hat{I}}_j$ ($j{>}0$) denote
the output image of the $j$-th SCN upscaled by a total of ${\times}s^j$ times.
Each $\mathbf{\hat{I}}_j$ can be compared with its associated ground truth image $\mathbf{I}_j$
according to the MSE cost, leading to a multi-scale objective function:
\begin{equation}
\label{eq:objcrnn}
    \min\limits_{\{\mathbf{\Theta}_j\}} \sum_i\sum_j
        {\left\|SCN(\mathbf{\hat{I}}_{j-1}^{(i)}{\uparrow}s; \mathbf{\Theta}_j) - \mathbf{I}_j^{(i)} \right\|^2_2},
\end{equation}
where $i$ denotes the data index, and $j$ denotes the SCN index.
$\mathbf{I}{\uparrow}s$ is the bicubic interpolated image of $\mathbf{I}$ by a factor of $s$.
This multi-scale objective function makes full use of the supervision information in all scales,
sharing a similar idea as heterogeneous networks \cite{Lee14deeply,changshiyu2015KDD}.
All the layer parameters $\{\mathbf{\Theta}_j\}$ in \eqref{eq:objcrnn} could be optimized
from end to end by back-propagation.
We use a greedy algorithm here to train each SCN sequentially from the beginning of the cascade
so that we do not need to care about the gradient of bicubic layers.
Applying back-propagation through a bicubic layer or its trainable surrogate will be considered in future work.


\section{Implementation Details}
\label{sec:exp_imp}

We determine the number of nodes in each layer of our SCN mainly
according to the corresponding settings used in sparse coding \cite{Yang12coupled}.
Unless otherwise stated, we use input LR patch size $s_y{=}9$, LR feature dimension $m_y{=}100$,
dictionary size $n{=}128$, output HR patch size $s_x{=}5$, and patch aggregation filter size $s_g{=}5$.
All the convolution layers have a stride of $1$.
Each LR patch $\y$ is normalized by its mean and variance, and the same mean and variance are used to restore
the final HR patch $\x$.
We crop $56{\times}56$ regions from each image to obtain fixed-sized input samples to the network,
which produces outputs of size $44{\times}44$.

To reduce the number of parameters, we implement the LR patch extraction layer $\mathbf{H}$ as the combination of two layers:
the first layer has $4$ trainable filters each of which is shifted to $25$ fixed positions by the second layer.
Similarly, the patch combination layer $\mathbf{G}$ is also split into a fixed layer which aligns pixels in overlapping patches
and a trainable layer whose weights are used to combine overlapping pixels.
In this way, the number of parameters in these two layers are reduced by more than an order, and there is no
observable loss in performance.

We employ a standard stochastic gradient descent algorithm to train our networks with mini-batch size of $64$.
Based on the understanding of each layer's role in sparse coding, we use Harr-like gradient filters to initialize layer $\mathbf{H}$,
and use uniform weights to initialize layer $\mathbf{G}$.
All the remaining three linear layers are related to the dictionary pair $(\D_x, \D_y)$ in sparse coding.
To initialize them, we first randomly set $\D_x$ and $\D_y$ with Gaussian noise,
and then find the corresponding layer weights as in ISTA \cite{Daubechies04}:
\begin{equation}
\label{eq:winit}
 \mathbf{w}_{1} = C \cdot \D_y^T, \;
 \mathbf{w}_{2} = \mathbf{I} - \D_y^T \D_y, \;
 \mathbf{w}_{3} = (CL)^{-1}\cdot\D_x
\end{equation}
where $\mathbf{w}_1$, $\mathbf{w}_2$ and $\mathbf{w}_3$ denote the weights of
the three subsequent layers after layer $\mathbf{H}$. $L$ is the upper bound on the largest eigenvalue
of $\D_y^T \D_y$, and $C$ is the threshold value before normalization. We empirically set
$L{=}C{=}5$.

The proposed models are all trained using the CUDA ConvNet package \cite{Krizhevsky12NIPS}
on a workstation with 12 Intel Xeon 2.67GHz CPUs and 1 GTX680 GPU.
Training a SCN usually takes less than one day.
Note that this package is customized for classification networks, and its efficiency can be further optimized
for our SCN model.

In testing, to make the entire image covered by output samples, we crop input samples with overlap
and extend the boundary of original image by reflection.
Note we shave the image border in the same way as \cite{Dong14ECCV} for objective evaluations
to ensure fair comparison.
Only the luminance channel is processed with our method, and bicubic interpolation is applied to
the chrominance channels.
To achieve arbitrary upscaling factors using CSCN, we upscale an image by $\times2$ times repeatedly
until it is at least as large as the desired size.
Then a bicubic interpolation is used to downscale it to the target resolution if necessary.

When reporting our best results in Sec.~\ref{sec:exp_cmp}, we also use the multi-view testing strategy commonly
employed in image classification. For patch-based image SR, multi-view testing is implicitly used when predictions from
multiple overlapping patches are averaged.
Here, besides sampling overlapping patches, we also add more views by flipping and transposing the patch.
Such strategy is found to improve SR performance for general algorithms at the sheer cost of computation.

\section{Experiments}
\label{sec:exp}

We evaluate and compare the performance of our models using the same data and protocols as in \cite{Timofte13ICCV},
which are commonly adopted in SR literature.
All our models are learned from a training set with 91 images, and tested on
Set5 \cite{Bevilacqua12BMVC}, Set14 \cite{Zeyde12Single} and BSD100 \cite{Martin01BSD} which contain 5, 14 and 100 images respectively.
We have also trained on a different larger data set, and observe little performance change (less than 0.1dB).
The original images are downsized by bicubic interpolation to generate LR-HR image pairs
for both training and evaluation. The training data are augmented with translation, rotation and scaling.

\subsection{Algorithm Analysis}

\begin{figure}[t]
\small
\center
    \includegraphics[height=0.1\linewidth,clip]{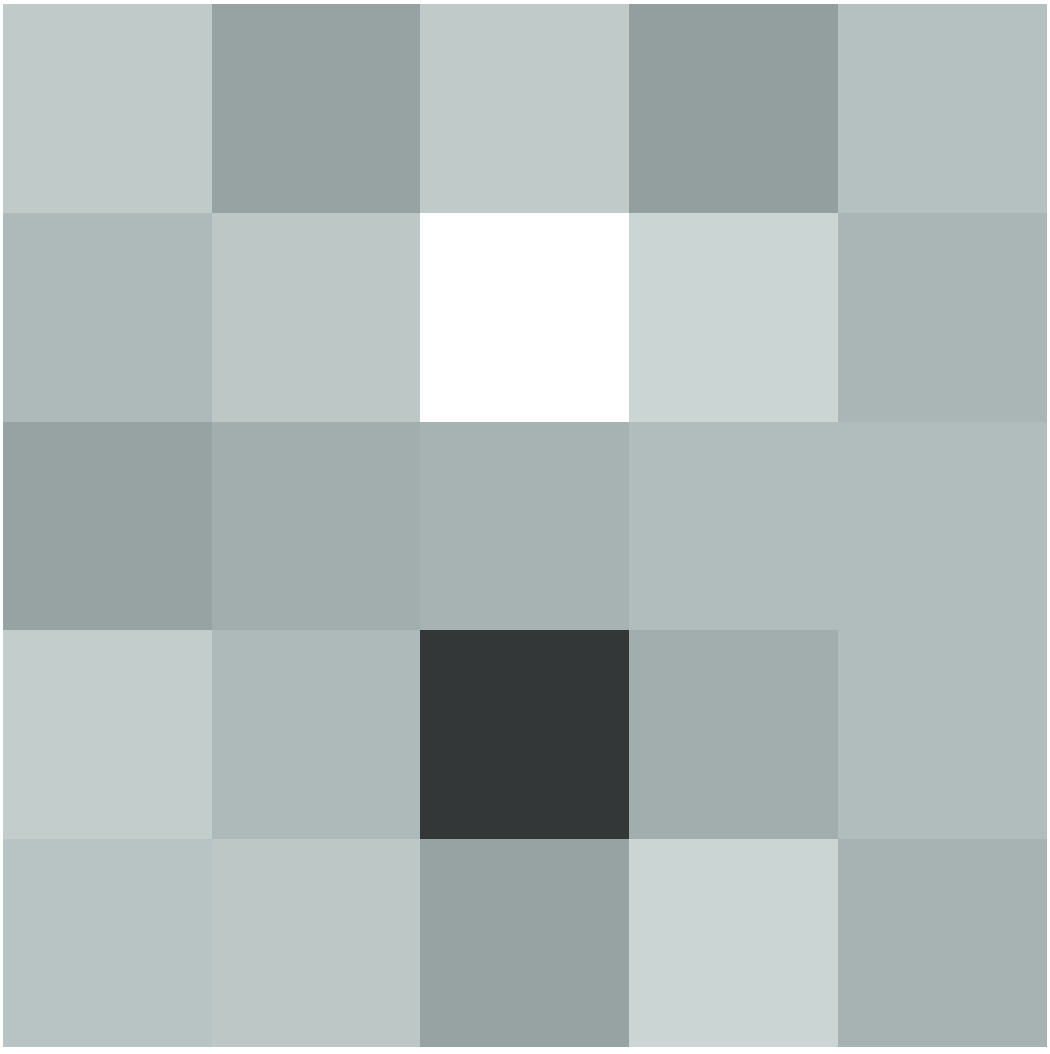}
    \includegraphics[height=0.1\linewidth,clip]{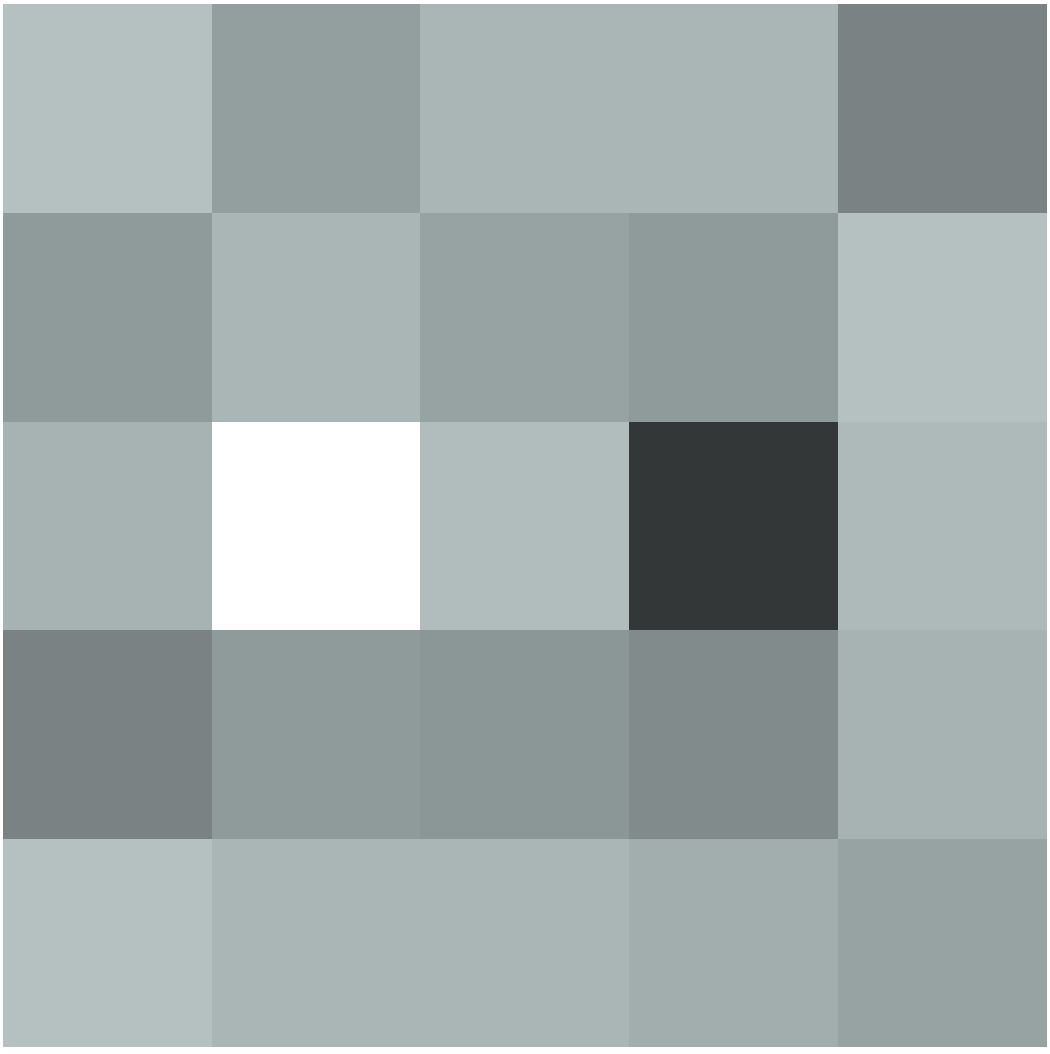}
    \includegraphics[height=0.1\linewidth,clip]{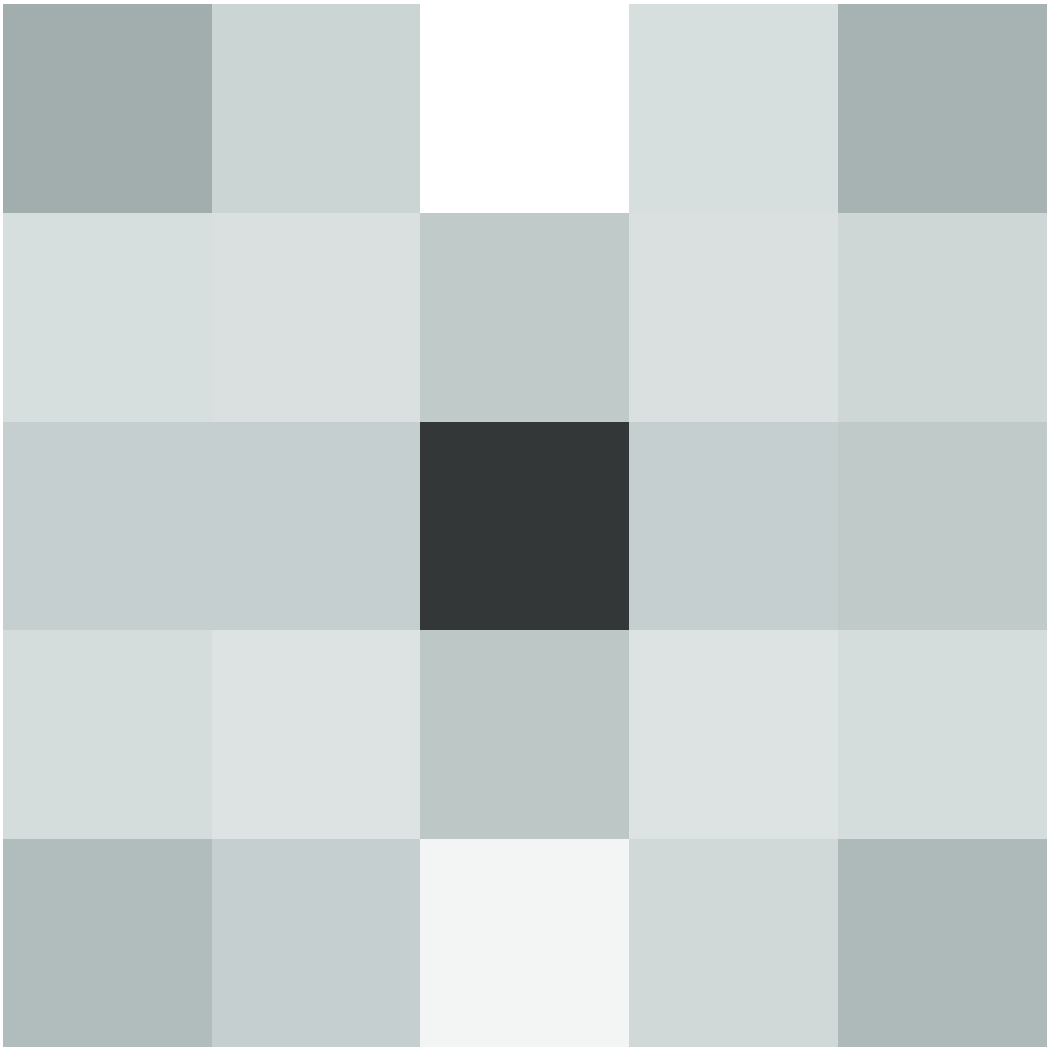}
    \includegraphics[height=0.1\linewidth,clip]{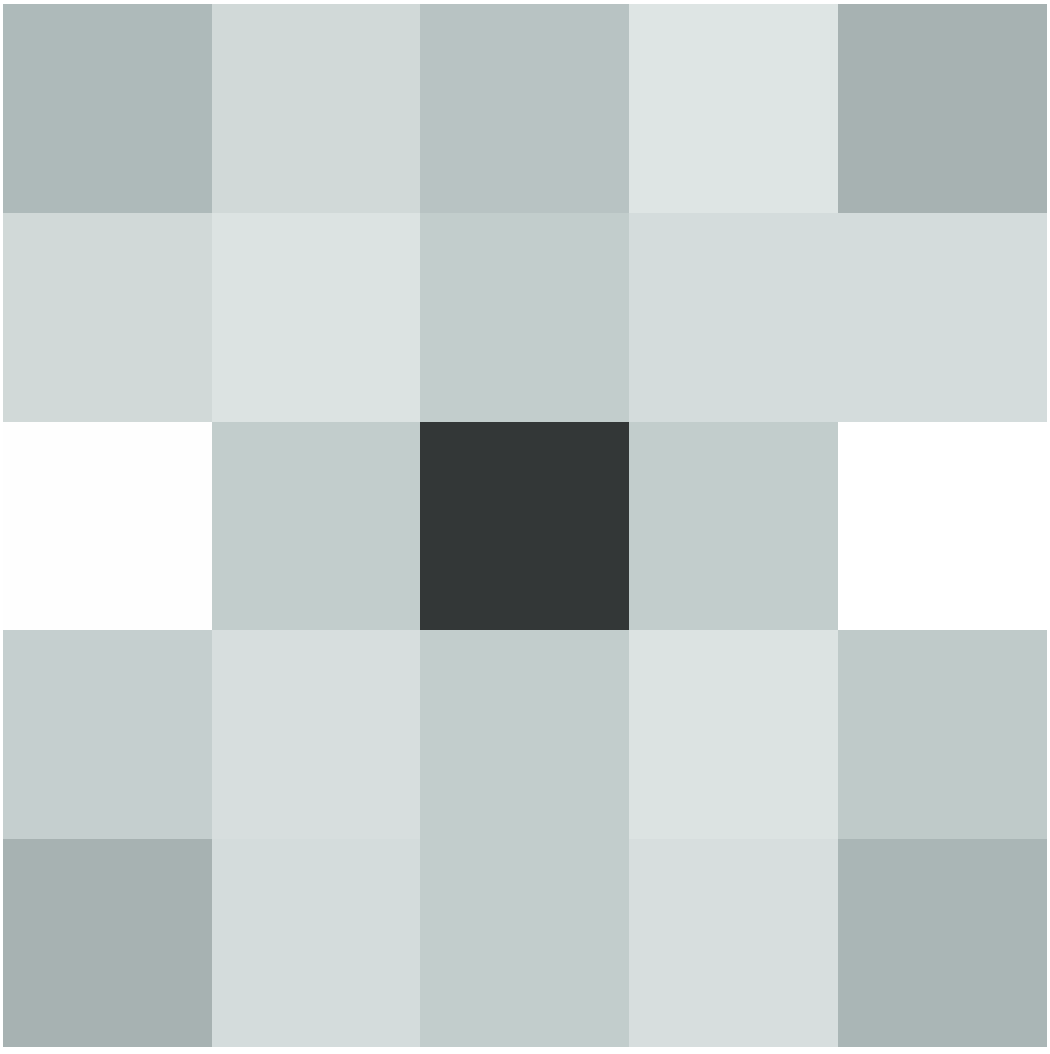}
\caption{The four learned filters in the first layer $\mathbf{H}$.}
\label{fig:conv1}
\end{figure}

We first visualize the four filters learned in the first layer $\mathbf{H}$ in Fig.~\ref{fig:conv1}.
The filter patterns do not change much from the initial first and second order gradient operators.
Some additional small coefficients are introduced in a highly structured form that capture
richer high frequency details.

\begin{figure}[t]
\small
\center
    \includegraphics[width=0.75\linewidth,clip]{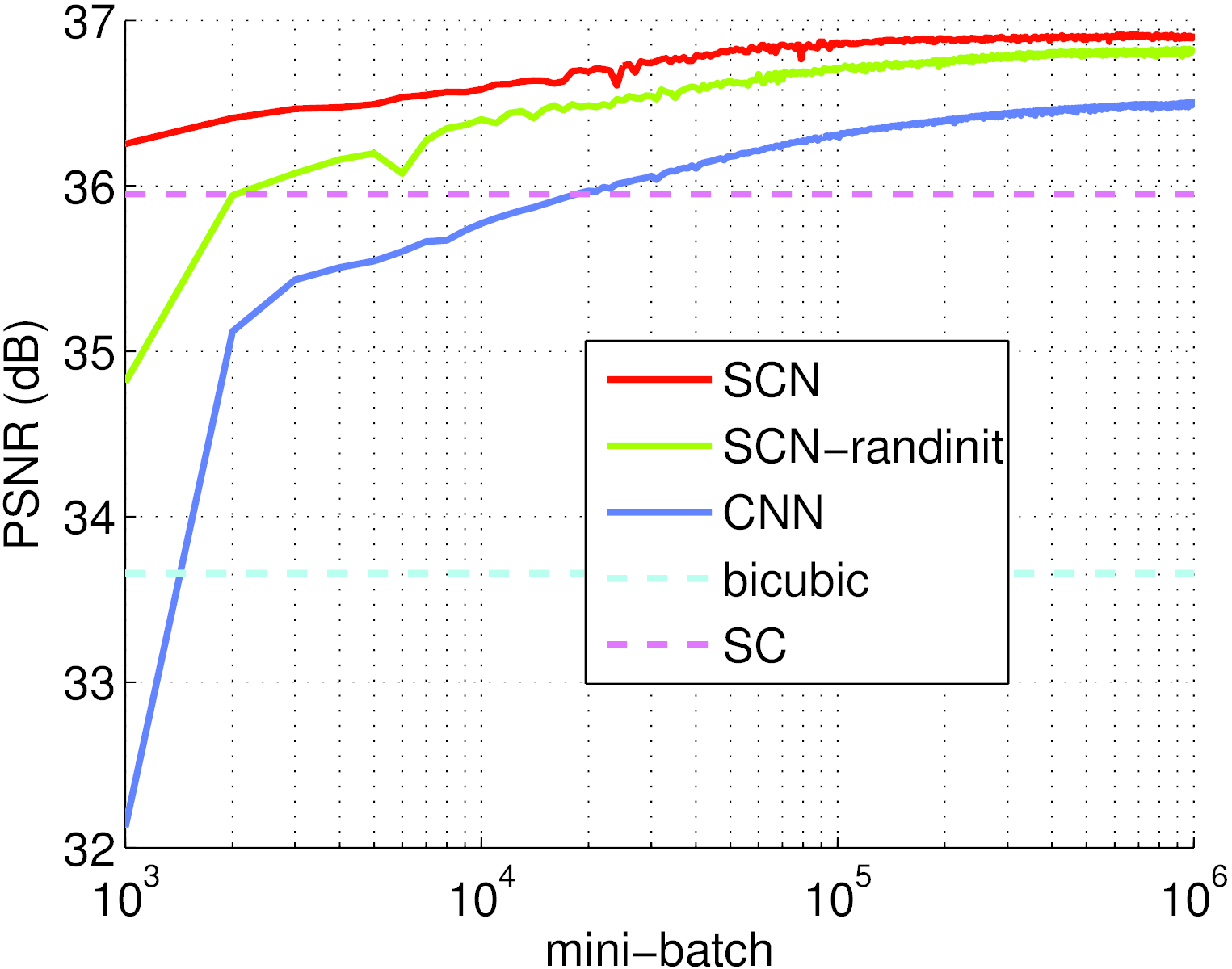}
\caption{The PSNR change for $\times2$ SR on Set5 during training using different methods: SCN; SCN with random initialization; CNN.
        The horizontal dash lines show the benchmarks of bicubic interpolation and sparse coding (SC).}
\label{fig:traincurve}
\end{figure}

The performance of several networks during training is measured on Set5 in Fig.~\ref{fig:traincurve}.
Our SCN improves significantly over sparse coding (SC) \cite{Yang12coupled}, as it leverages data more effectively
with end-to-end training.
The SCN initialized according to \eqref{eq:winit} can converge faster and better than the same model with random initialization,
which indicates that the understanding of SCN based on sparse coding can help its optimization.
We also train a CNN model \cite{Dong14ECCV} of the same size as SCN, but find its convergence speed much slower.
It is reported in \cite{Dong14ECCV} that training a CNN takes $8{\times}10^8$ back-propagations (equivalent to $12.5{\times}10^6$ mini-batches here).
To achieve the same performance as CNN, our SCN requires less than 1\% back-propagations.

\begin{figure}[t]
\small
\center
    \includegraphics[width=0.85\linewidth,clip]{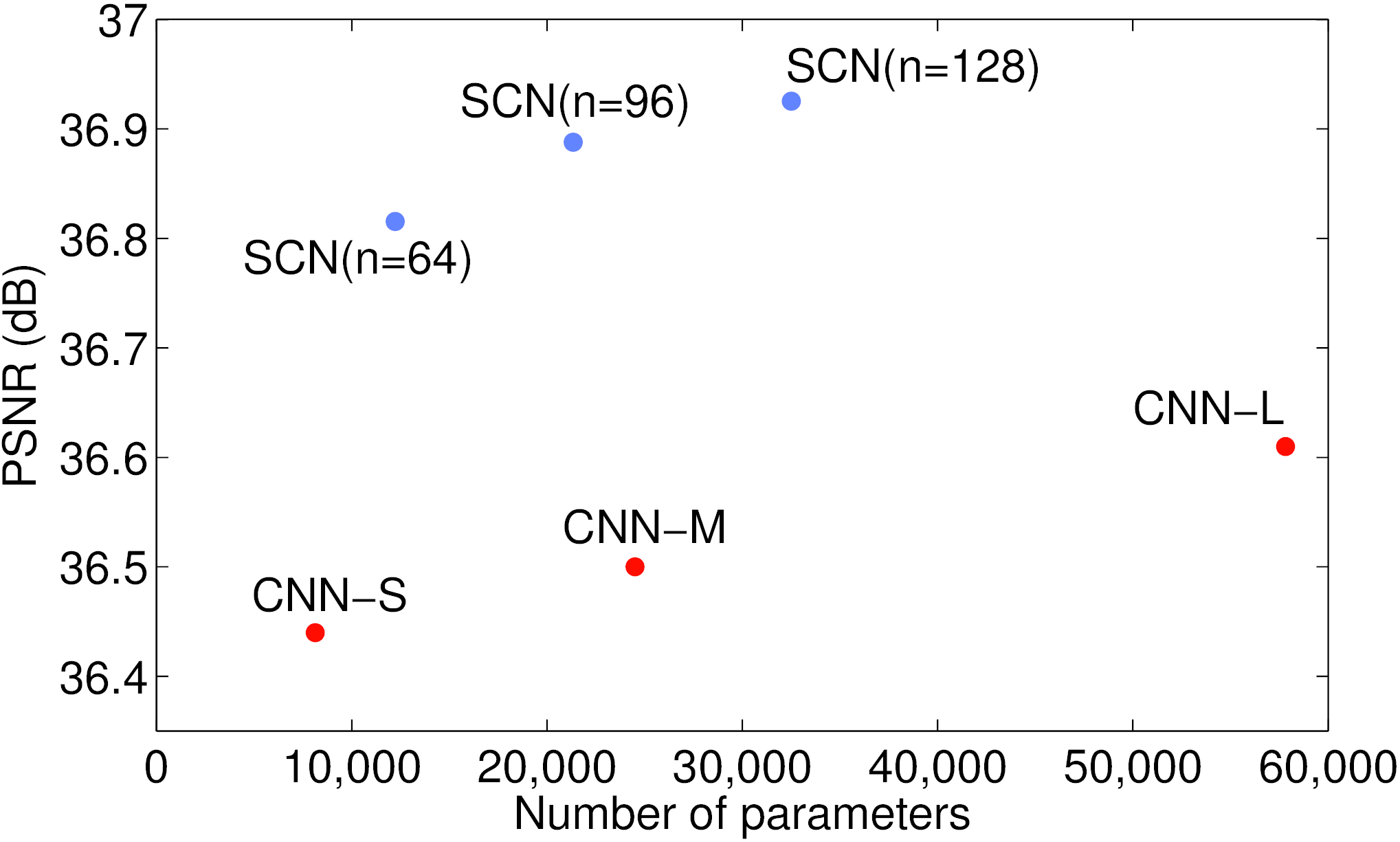}
\caption{PSNR for $\times2$ SR on Set5 using SCN and CNN with various network sizes.}
\label{fig:paramcurve}
\end{figure}


The network size of SCN is mainly determined by the dictionary size $n$. Besides the default value $n{=}128$, we have tried other
sizes and plot their performance versus the number of network parameters in Fig.~\ref{fig:paramcurve}.
The PSNR of SCN does not drop too much as $n$ decreases from $128$ to $64$,
but the model size and computation time can be reduced significantly.
Fig.~\ref{fig:paramcurve} also shows the performance of CNN with various sizes.
Our smallest SCN can achieve higher PSNR than the largest model (CNN-L) in \cite{Dong14arxiv} while only using
about 20\% parameters.

Different numbers of recurrent stages $k$ have been tested for SCN, and we find increasing $k$ from $1$ to $3$ only improves
performance by less than 0.1dB. As a tradeoff between speed and accuracy, we use $k{=}1$ throughout the paper.

\begin{table}[t]
\centering
\caption{PSNR of different network cascading schemes on Set5, evaluated for different scaling factors in each column.}
\label{tab:netstruct}
\vspace{1pt}
\begin{tabular}{r||cccc}
\hline
upscale factor  & $\times1.5$	& $\times2$	& $\times3$ & $\times4$ \\
\hline
SCN$\times1.5$ & 40.14 &	36.41 &	30.33 &	29.02 \\
SCN$\times2$   & \textbf{40.15} &	\textbf{36.93} &	32.99 &	30.70 \\
SCN$\times3$   & 39.88 &	36.76 &	32.87 &	30.63 \\
SCN$\times4$   & 39.69 &	36.54 &	32.76 &	30.55 \\
\hline
CSCN	& \textbf{40.15} &	\textbf{36.93} &	\textbf{33.10} &	\textbf{30.86}  \\
\hline
\end{tabular}
\end{table}

In Table~\ref{tab:netstruct}, different network cascade structures (in each row) are compared
at different scaling factors (in each column).
SCN$\times{a}$ denotes the simple cascade of SCN with fixed scaling factor $a$,
where an individually trained SCN is applied one or more times for scaling factors other than $a$.
It is observed that SCN$\times2$
can perform as well as the scale-specific model for small scaling factor (1.5),
and much better for large scaling factors (3 and 4).
Note that the cascade of SCN$\times1.5$ does not lead to good results since
artifacts quickly get amplified through many repetitive upscalings.
Therefore, we use SCN$\times2$ as the default building block for CSCN,
and drop the notation $\times2$ when there is no ambiguity.
The last row in Table~\ref{tab:netstruct} shows that a CSCN trained using
the multi-scale objective in \eqref{eq:objcrnn} can further improve the SR results
for scaling factors 3 and 4, as the second SCN in the cascade is trained to be
robust to the artifacts generated by the first one.

\subsection{Comparison with State of the Arts}
\label{sec:exp_cmp}

We compare the proposed CSCN with other recent SR methods on all the images in Set5, Set14 and BSD100
for different upscaling factors.
Table~\ref{tab:set5} shows the PSNR and structural similarity (SSIM) \cite{Wang04SSIM} for
adjusted anchored neighborhood regression (A+) \cite{Timofte14A+}, CNN \cite{Dong14ECCV},
CNN trained with larger model size and more data (CNN-L) \cite{Dong14arxiv}, the proposed CSCN,
and CSCN with our multi-view testing (CSCN-MV). We do not list other methods \cite{Yang12coupled,Timofte13ICCV,Zeyde12Single,Kim10PAMI,Huang15Trans}
whose performance is worse than A+ or CNN-L.

It can be seen from Table~\ref{tab:set5} that CSCN performs consistently better than all previous methods in both PSNR and SSIM,
and with multi-view testing the results can be further improved.
CNN-L improves over CNN by increasing model parameters and training data.
However, it is still not as good as CSCN which is trained with a much smaller size and on a much smaller data set.
Clearly, the better model structure of CSCN makes it less dependent on model capacity and training data
in improving performance.
Our models are generally more advantageous for large scaling factors due to the cascade structure.

\begin{table*}[t]
\centering
\caption{PSNR (SSIM) comparison on three test data sets among different methods.
        \textcolor{red}{Red} indicates the best and \textcolor{blue}{blue} indicates the second best performance.
        The performance gain of our best model over all the others' best is shown in the last row.}
\label{tab:set5}
\vspace{1pt}
\small
\begin{tabular}{r||c|c|c||c|c|c||c|c|c}
\hline
Data Set      & \multicolumn{3}{c||}{Set5}	& \multicolumn{3}{c||}{Set14}	& \multicolumn{3}{c}{BSD100} \\
\hline
Upscaling &	$\times2$ &	$\times3$ & $\times4$ &
        	$\times2$ &	$\times3$ & $\times4$ &
            $\times2$ &	$\times3$ & $\times4$ \\
\hline
\hline
\multirow{2}{*}{A+ \cite{Timofte14A+}} &
36.55 & 32.59 & 30.29 & 32.28 & 29.13 & 27.33 & 30.78 & 28.18 & 26.77 \\
& (0.9544) & (0.9088) & (0.8603) & (0.9056) & (0.8188) & (0.7491) & (0.8773) & (0.7808) & (0.7085) \\
\hline
\multirow{2}{*}{CNN \cite{Dong14ECCV}} &
36.34 & 32.39 & 30.09 & 32.18 & 29.00 & 27.20 & 31.11 & 28.20 & 26.70 \\
& (0.9521) & (0.9033) & (0.8530) & (0.9039) & (0.8145) & (0.7413) & (0.8835) & (0.7794) & (0.7018) \\
\hline
\multirow{2}{*}{CNN-L \cite{Dong14arxiv}} &
36.66 & 32.75 & 30.49 & 32.45 & 29.30 & 27.50 & 31.36 & 28.41 & 26.90 \\
& (0.9542) & (0.9090) & (0.8628) & (0.9067) & (0.8215) & (0.7513) & (0.8879) & (0.7863) & (0.7103) \\
\hline
\multirow{2}{*}{CSCN} &
\textcolor{blue}{36.93} & \textcolor{blue}{33.10} & \textcolor{blue}{30.86} & \textcolor{blue}{32.56} & \textcolor{blue}{29.41} &
\textcolor{blue}{27.64} & \textcolor{blue}{31.40} & \textcolor{blue}{28.50} & \textcolor{blue}{27.03} \\
& \textcolor{blue}{(0.9552)} & \textcolor{blue}{(0.9144)} & \textcolor{blue}{(0.8732)} & \textcolor{blue}{(0.9074)} &
\textcolor{blue}{(0.8238)} & \textcolor{blue}{(0.7578)} & \textcolor{blue}{(0.8884)} & \textcolor{blue}{(0.7885)} & \textcolor{blue}{(0.7161)} \\
\hline
\multirow{2}{*}{CSCN-MV} &
\textcolor{red}{37.14} & \textcolor{red}{33.26} & \textcolor{red}{31.04} & \textcolor{red}{32.71} & \textcolor{red}{29.55} &
\textcolor{red}{27.76} & \textcolor{red}{31.54} & \textcolor{red}{28.58} & \textcolor{red}{27.11} \\
& \textcolor{red}{(0.9567)} & \textcolor{red}{(0.9167)} & \textcolor{red}{(0.8775)} & \textcolor{red}{(0.9095)} &
\textcolor{red}{(0.8271)} & \textcolor{red}{(0.7620)} & \textcolor{red}{(0.8908)} & \textcolor{red}{(0.7910)} & \textcolor{red}{(0.7191)} \\
\hline
\hline
Our &       0.48 & 0.51 & 0.55 & 0.26 & 0.25 & 0.26 & 0.18 & 0.17 & 0.21 \\
Improvement & (0.0023) & (0.0077) & (0.0147) & (0.0028) & (0.0056) & (0.0107) & (0.0029) & (0.0047) & (0.0088) \\
\hline
\end{tabular}
\end{table*}

\begin{figure*}[t]
\small
\center
    \begin{tabular}{p{2mm}@{\hskip 2mm}c@{\hskip 1mm}c@{\hskip 1mm}c}
    \rotatebox{90}{\hspace{15mm} SC} &
	\includegraphics[height=43mm,angle=0,viewport=120 0 743 495,clip=true]{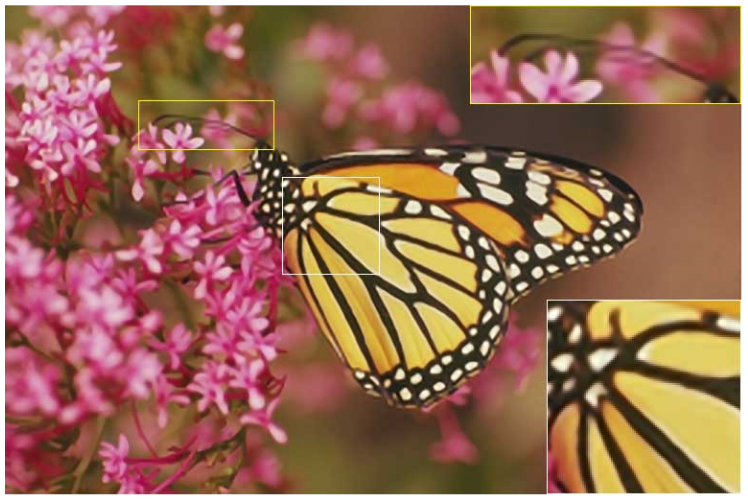} &
    \includegraphics[height=43mm,angle=0,viewport=20 0 703 469,clip=true]{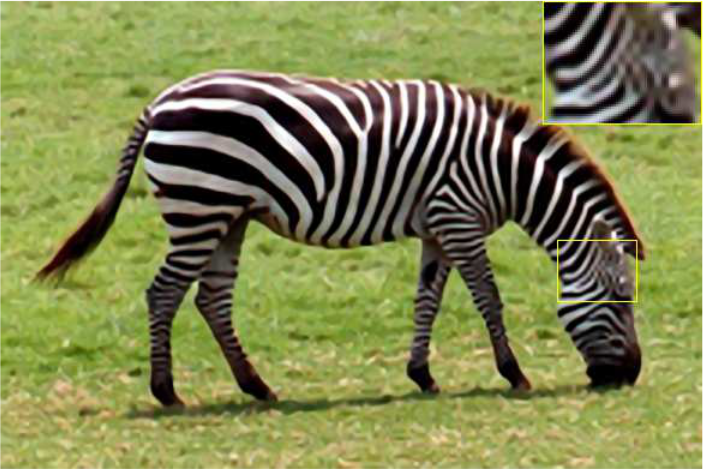} &
    \includegraphics[height=43mm,angle=0,viewport=10 105 245 351,clip]{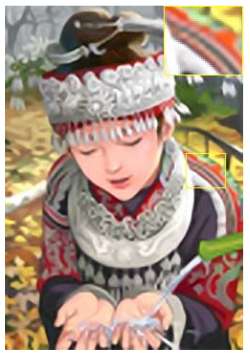} \\
    \rotatebox{90}{\hspace{14mm} CNN} &
	\includegraphics[height=43mm,angle=0,viewport=120 0 743 495,clip]{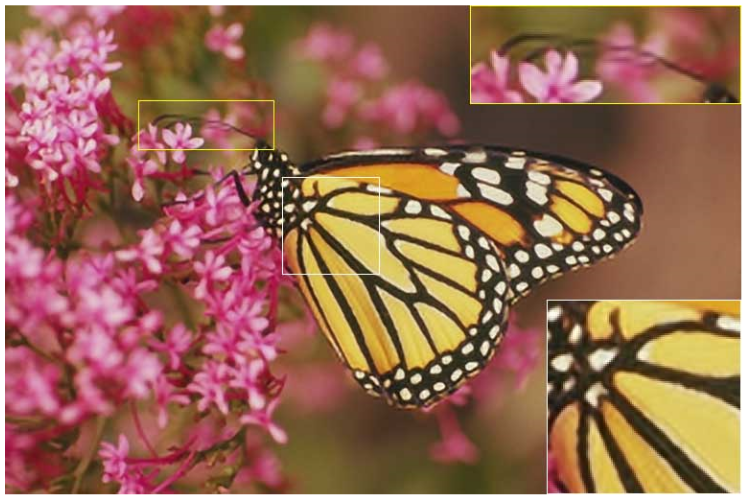} &
    \includegraphics[height=43mm,angle=0,viewport=20 0 703 469,clip]{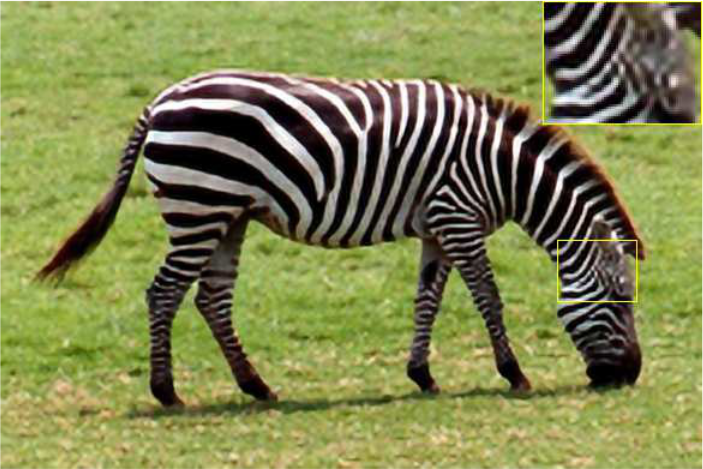} &
    \includegraphics[height=43mm,angle=0,viewport=10 105 245 351,clip]{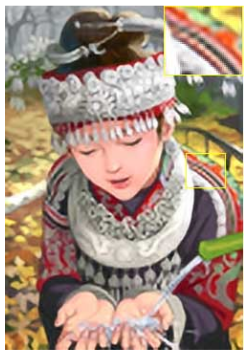} \\
    \rotatebox{90}{\hspace{14mm} CSCN} &
	\includegraphics[height=43mm,angle=0,viewport=120 0 743 495,clip]{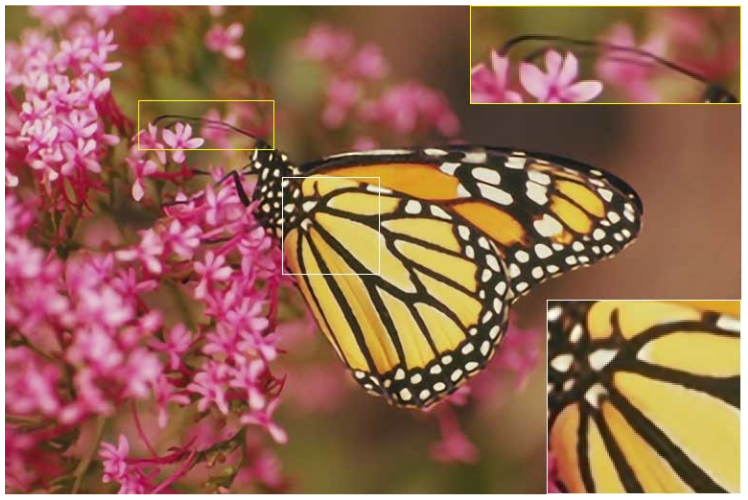} &
    \includegraphics[height=43mm,angle=0,viewport=20 0 703 469,clip]{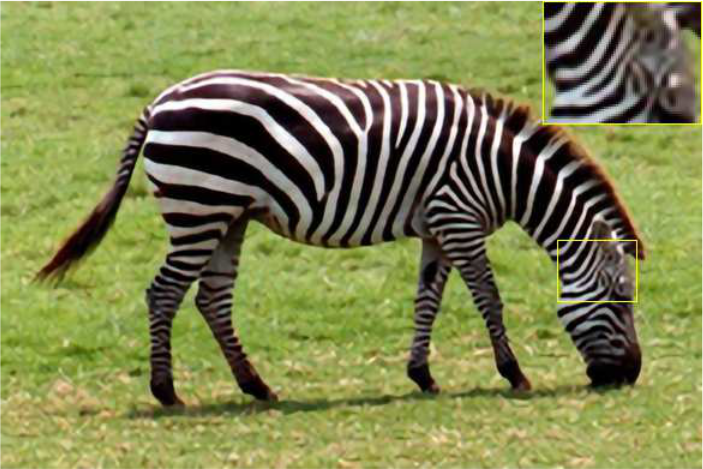} &
    \includegraphics[height=43mm,angle=0,viewport=10 105 245 351,clip]{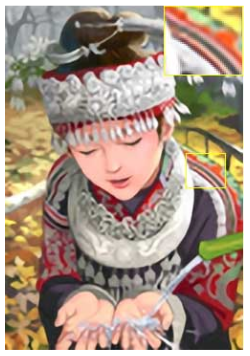} \\
    \end{tabular}
\caption{SR results given by SC \cite{Yang12coupled} (first row), CNN \cite{Dong14ECCV} (second row) and our CSCN (third row).
            Images from left to right: the ``monarch'' image upscaled by $\times3$;
            the ``zebra'' image upscaled by $\times3$; the ``comic'' image upscaled by $\times3$.}
\label{fig:monarch}
\end{figure*}


\begin{figure*}[t]
\small
\center
    \begin{tabular}{c@{\hskip 1mm}c@{\hskip 1mm}c}
    \includegraphics[height=0.32\linewidth,angle=-90,bb=160 140 600 754,clip]{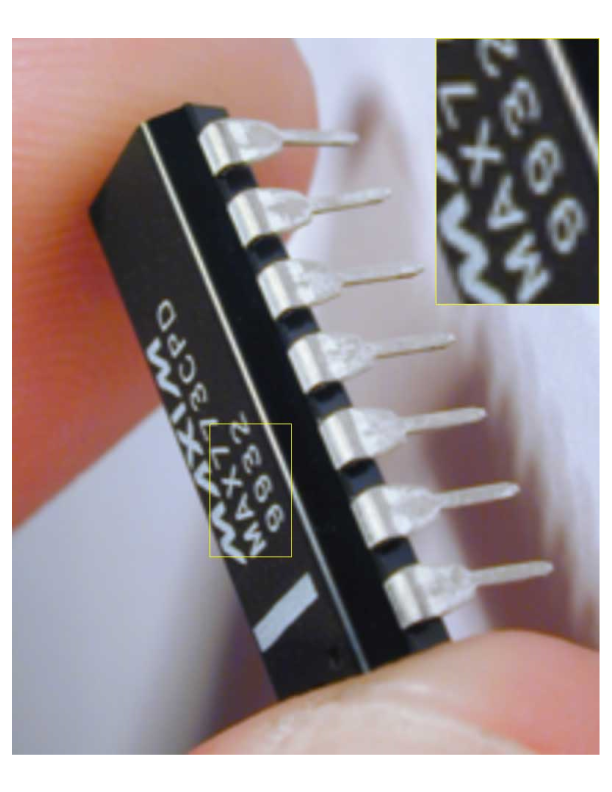} &
    \includegraphics[height=0.32\linewidth,angle=-90,bb=160 140 600 754,clip]{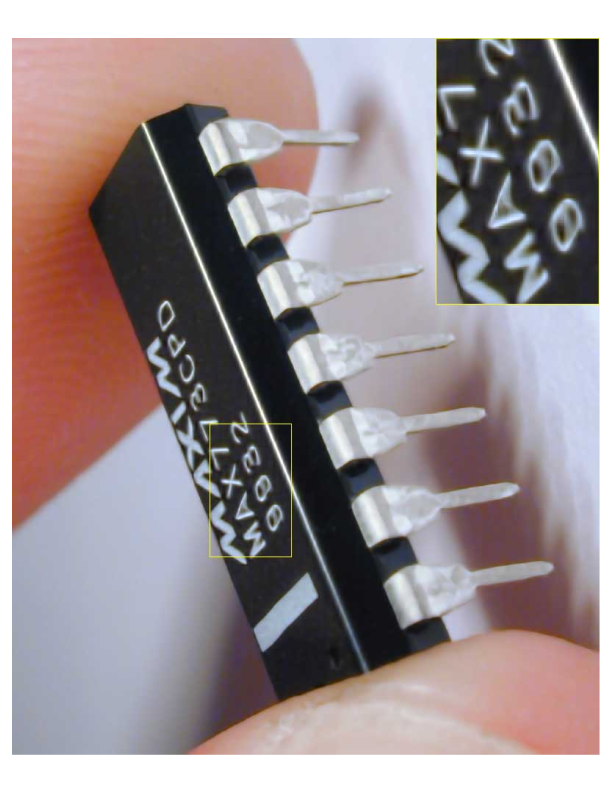} &
    \includegraphics[height=0.32\linewidth,angle=-90,bb=160 140 600 754,clip]{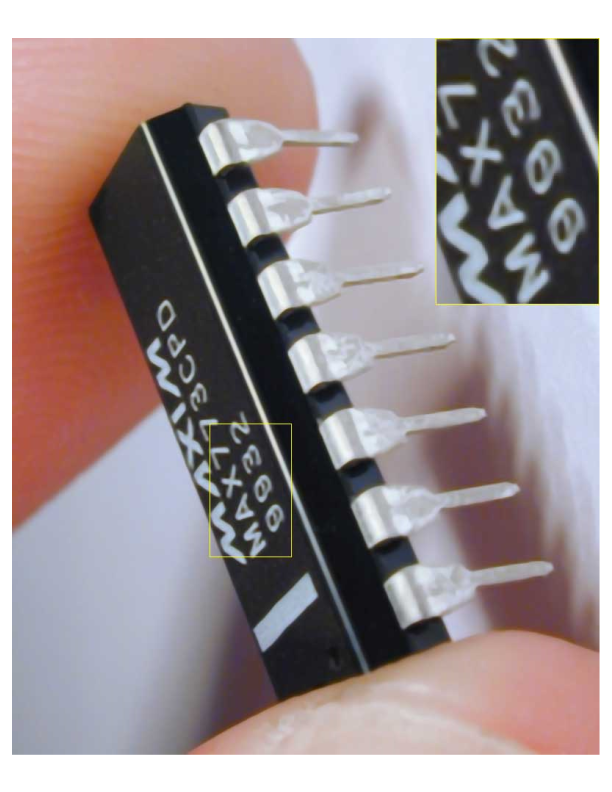} \\
    (a) bicubic & (b) SE \cite{Freedman11TG} & (c) SC \cite{Yang12coupled} \vspace{-2mm} \\
    \includegraphics[height=0.32\linewidth,bb=160 140 600 754,angle=-90,clip]{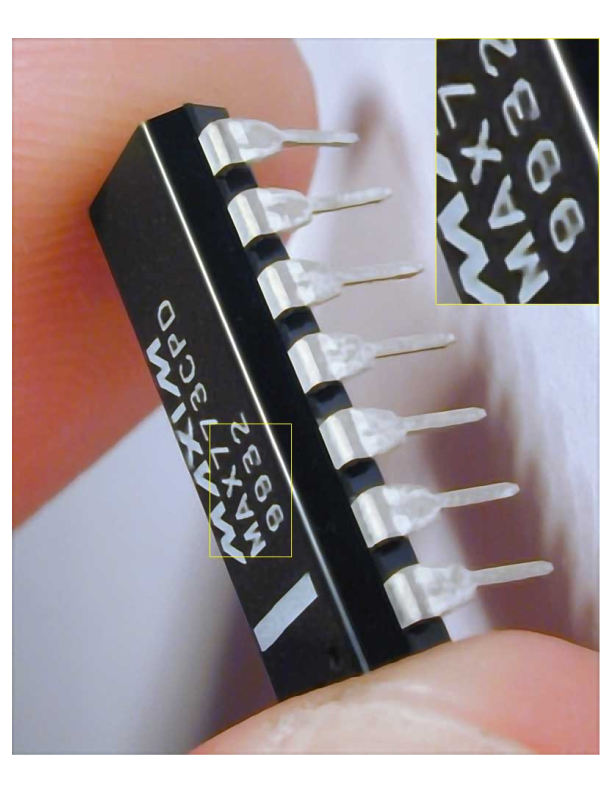} &
    \includegraphics[height=0.32\linewidth,bb=160 140 600 754,angle=-90,clip]{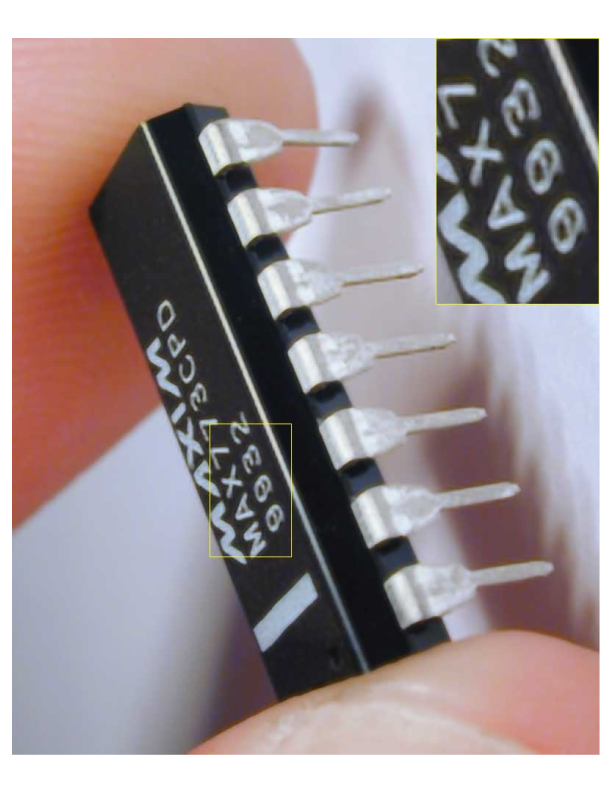} &
    \includegraphics[height=0.32\linewidth,bb=160 140 600 754,angle=-90,clip]{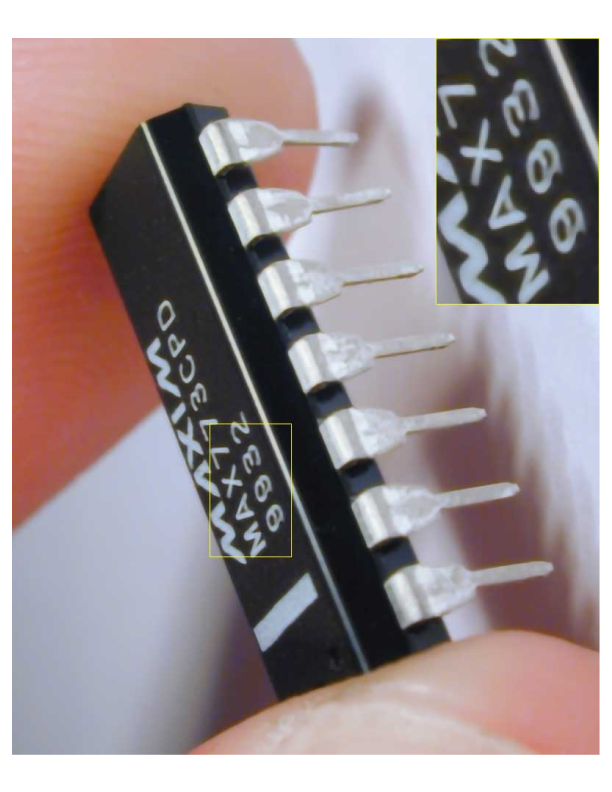} \\
    (d) DNC \cite{Cui14ECCV} & (e) CNN \cite{Dong14ECCV} & (f) CSCN
    \end{tabular}
\caption{The ``chip'' image upscaled by $\times4$ times using different methods.}
\vspace{-4mm}
\label{fig:chip}
\end{figure*}

The visual qualities of the SR results generated by sparse coding (SC) \cite{Yang12coupled}, CNN and CSCN
are compared in Fig.~\ref{fig:monarch}.
Our approach produces image patterns with shaper boundaries and richer textures,
and is free of the ringing artifacts observable in the other two methods.

Fig.~\ref{fig:chip} shows the SR results on the ``chip'' image compared among more methods including
the self-example based method (SE) \cite{Freedman11TG} and the deep network cascade (DNC) \cite{Cui14ECCV}.
SE and DNC can generate very sharp edges on this image, but also introduce artifacts and blurs
on corners and fine structures due to the lack of self-similar patches.
On the contrary, the CSCN method recovers all the structures of the characters without any distortion.

We also compare CSCN with other sparse coding extensions \cite{lu2012geometry,dong2011centralized,zhang2012single},
and consider the blurring effect introduced in downscaling.
A PSNR gain of 0.3$\sim$1.6dB is achieved by CSCN in general.
Experiment details and source codes are available online\footnote{\url{www.ifp.illinois.edu/~dingliu2/iccv15}}.

\subsection{Subjective Evaluation}

\begin{figure}[t]
\small
\center
    \includegraphics[width=0.75\linewidth,clip]{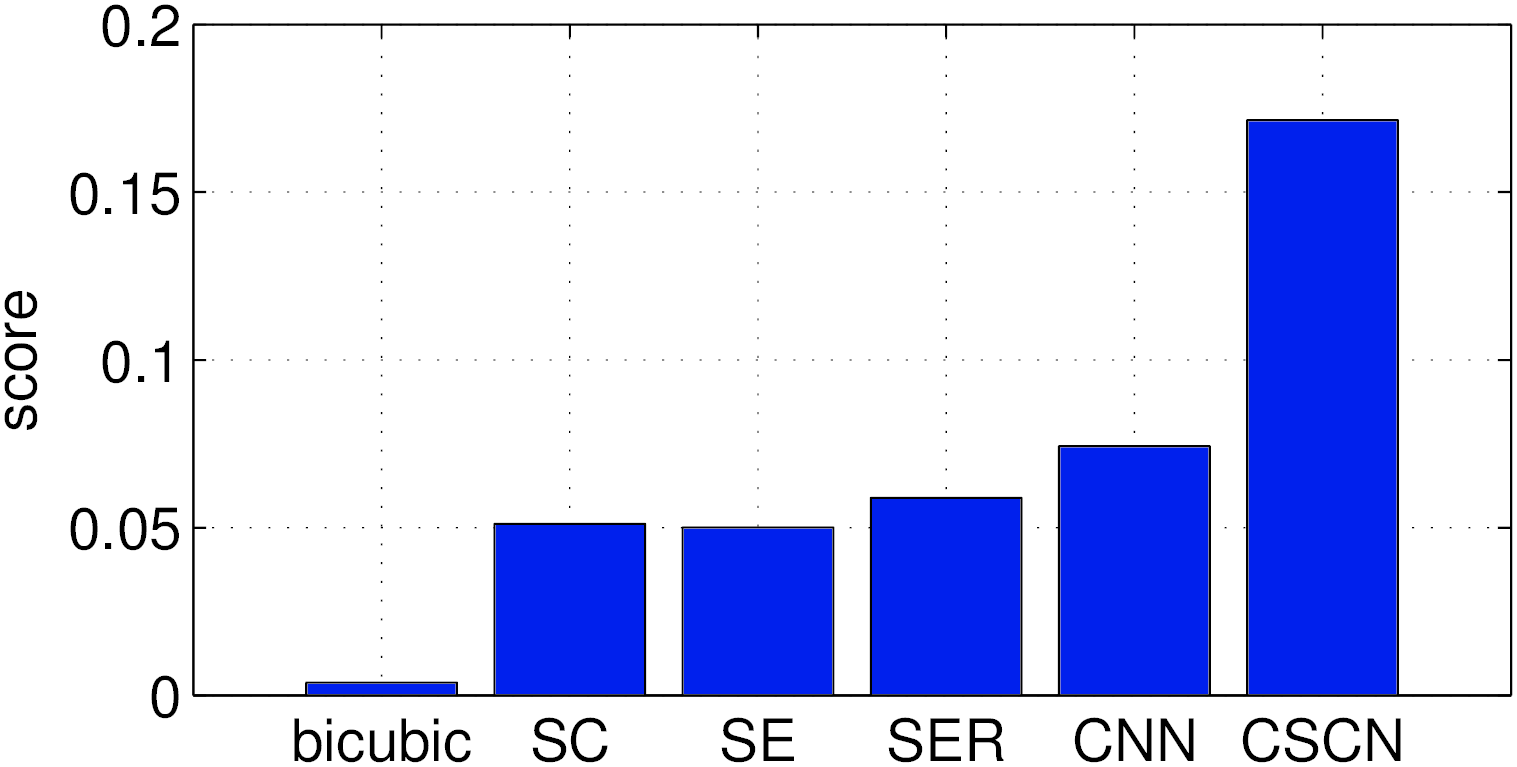}
\caption{Subjective SR quality scores for different methods including bicubic, SC \cite{Yang12coupled},
        SE \cite{Freedman11TG}, SER \cite{yang13Inplace}, CNN \cite{Dong14ECCV} and the proposed CSCN.
        The score for ground truth result is 1.}
\label{fig:subeval}
\end{figure}

We conducted a subjective evaluation of SR results for several methods
including bicubic, SC \cite{Yang12coupled}, SE \cite{Freedman11TG}, self-example regression (SER) \cite{yang13Inplace},
CNN \cite{Dong14ECCV} and CSCN.
Ground truth HR images are also included when they are available as references.
Each of the participants in the evaluation is shown a set of HR image pairs,
which are upscaled from the same LR images using two randomly selected methods.
For each pair, the subject needs to decide which one is better in terms of perceptual quality.

We have a total of 270 participants giving 720 pairwise comparisons over 6 images with different scaling factors.
Not every participant completed all the comparisons but their partial responses are still useful.
All the evaluation results can be summarized into a $7{\times}7$ winning matrix for 7 methods (including ground truth),
based on which we fit a Bradley-Terry \cite{Bradley52rank} model to estimate
the subjective score for each method so that they can be ranked.

Fig.~\ref{fig:subeval} shows the estimated scores for the 6 SR methods in our evaluation,
with the score for ground truth method normalized to 1.
As expected, all the SR methods have much lower scores than ground truth, showing the great challenge in SR problem.
The bicubic interpolation is significantly worse than other SR methods.
The proposed CSCN method outperforms other previous state-of-the-art methods by a large margin,
demonstrating its superior visual quality.
It should be noted that the visual difference between some image pairs is very subtle.
Nevertheless, the human subjects are able to perceive such difference when seeing the two images side by side,
and therefore make consistent ratings.
The CNN model becomes less competitive in the subjective evaluation than it is in PSNR comparison.
This indicates that the visually appealing image appearance produced by CSCN should be attributed to
the regularization from sparse representation,
which can not be easily learned by merely minimizing reconstruction error as in CNN.



\section{Conclusions}
\label{sec:cncl}

We propose a new model for image SR by combining the strengths of sparse coding and deep network,
and make considerable improvement over existing deep and shallow SR models both quantitatively and qualitatively.
Besides producing good SR results, the domain knowledge in the form of sparse coding can also benefit
training speed and model compactness.
Furthermore, we propose a cascaded network for better flexibility in scaling factors as well as
more robustness to artifacts.

In future work, we will apply the SCN model to other problems where sparse coding can be useful.
The interaction between deep networks for low-level and high-level vision tasks will also be explored.

\vfill



{\small
\bibliographystyle{ieee}
\bibliography{ref}
}

\end{document}